\documentclass[conference]{IEEEtran}
\makeatletter
\def\ps@IEEEtitlepagestyle{%
  \def\@oddfoot{\mycopyrightnotice}%
  \def\@evenfoot{}%
}
\def\mycopyrightnotice{%
  {\footnotesize 978-1-5090-6004-7/17/\$31.00 \textcopyright 2017 IEEE\hfill}% <--- Change here
  \gdef\mycopyrightnotice{}% just in case
}

\usepackage{cite}
\ifCLASSINFOpdf
   \usepackage[pdftex]{graphicx}
\else
\fi

\usepackage{array}
\newcolumntype{P}[1]{>{\centering\arraybackslash}p{#1}}
\newcolumntype{M}[1]{>{\centering\arraybackslash}m{#1}}

\usepackage{fixltx2e}
\begin{document}

\title{Chord Angle Deviation using Tangent (CADT), an Efficient and Robust Contour-based Corner Detector}

%\author{\IEEEauthorblockN{Mohammad Asiful Hossain}
%\IEEEauthorblockA{Computer Science and Engineering Department\\
%%Bangladesh University of Engineering and Technology\\
%University of Asia Pacific \\
%Dhaka, Bangladesh\\
%Email: asif.hossain@uap-bd.edu}
%\and
%\IEEEauthorblockN{Abdul Kawsar Tushar}
%\IEEEauthorblockA{Computer Science and Engineering Department\\
%%Bangladesh University of Engineering and Technology\\
%University of Asia Pacific \\ 
%Dhaka, Bangladesh\\
%Email: tushar.kawsar@gmail.com}}


\author{\IEEEauthorblockN{Mohammad Asiful Hossain and
		Abdul Kawsar Tushar
	}
	\IEEEauthorblockA{Computer Science and Engineering Department, University of Asia Pacific, Dhaka, Bangladesh\\asif.hossain@uap-bd.edu, tushar.kawsar@gmail.com}
	%\IEEEauthorblockA{\IEEEauthorrefmark{2}Department of Electrical, Electronics, and Computer Engineering, University of Ulsan, Ulsan, Republic of Korea}
	%\IEEEauthorblockA{\IEEEauthorrefmark{3}Starfleet Academy, San Francisco, CA 96678 USA}
	%\IEEEauthorblockA{\IEEEauthorrefmark{4}Tyrell Inc., 123 Replicant Street, Los Angeles, CA 90210 USA}
	% <-this % stops an unwanted space
}


%Telephone: (800) 555--1212, Fax: (888) 555--1212}


\maketitle

\begin{abstract}
%\boldmath
%Corner detection has long been a pivotal process in numerous image processing and computer vision related applications. Among all the formerly published detectors described in this paper that are contour-based, 
%placeholder \newline
%placeholder \newline
%placeholder \newline
%placeholder \newline
%placeholder \newline
%placeholder \newline
%placeholder \newline
%placeholder \newline
%placeholder \newline
%placeholder \newline
Detection of corner is the most essential process in a large number of computer vision and image processing applications. We have mentioned a number of popular contour-based corner detectors  in our paper. Among all these detectors chord to triangular arm angle (CTAA) has been demonstrated as the most dominant corner detector in terms of average repeatability. We introduce a new effective method to calculate the value of curvature in this paper. By demonstrating experimental results, our proposed technique outperforms CTAA and other detectors mentioned in this paper. The results exhibit that our proposed method is simple yet efficient at finding out corners more accurately and reliably.
\end{abstract}


\textbf{Keywords - Image processing, Computer vision, Corner detection, Average repeatability, CTAA.}

% \fi
\IEEEpeerreviewmaketitle

\section{Introduction}
Feature detection of images has been a cornerstone of modern image processing, which has had a profound influence on various computer vision and image processing applications. From an image processing point of view, a feature is a piece of relevant image information with help of which various computational tasks can be solved which are related to a certain range of applications. Examples of these applications include edge detection, point of object detection, corner detection etc. A corner could be termed as location of an image edge where angle of the slope changes suddenly \cite{PE:PG:PG2011short:037-042}. This is the exact property that this paper aims to utilize for finding a better method of corner detection.

Among the famous chord-based corner detectors, Chord to Point Distance Accumulation (CPDA) technique is a well known one \cite{awrangjeb2012performance, awrangjeb2009fast, shui2013corner, awrangjeb2010comparative}. It uses a number of chords of varying lengths and then calculates a joint accumulated value of distance as the measure of curvature for a point. The probability of a point being a candidate corner is proportional to the value of the curvature. The benefit of CPDA is its relatively lower localization error. Nonetheless, when we consider using multiple chords being followed by a somewhat expensive process of refinement, they make CPDA quite an expensive process. A more recent example of corner detector is Chord to Triangular Arms Ratio (CTAR) which is also more effective and much simpler process \cite{PE:PG:PG2011short:037-042}. The CTAR method employs a simple triangulation theory to determine the value of curvature in each point, where the major advantage is limiting the number of chrods from multiple in CPDA to just one in CTAR. CTAR refinement process is also simpler in comparison to that of CPDA. This comparative study of corner detectors in \cite{awrangjeb2013performance} finds CTAR to be more effective and faster than CPDA in the field of average repeatability. However, CPDA has a lower localization error compared to that of CTAR. Chord to Triangular Arm Angle (CTAA) \cite{DBLP:conf/icsipa/HossainMS15} is another image corner detector which employs the method of using estimated corner angle as the value of curvature in order to find corners. \cite{DBLP:conf/icsipa/HossainMS15} demonstrates that CTAA performs better than CTAR in terms of average repeatability as well as localization error. Still, CPDA has a lower localization error compared to CTAA.

This paper proposes an algorithm which %performs better than CTAA in the field of average repeatability. CTAA uses the angle of corner which is estimated at one edge location as the value of curvature. Our method 
uses the concept of tangent along with a single chord to determine the corner points. Our experiments demonstrate that our proposed corner detector outperforms the CTAA method in the fields of average repeatability as our method uses simple mathematical operators and equations. We find an optimal angle value as well as chord length which give us better results. 

%The rest of this paper is organized as follows. Section II discusses the contributions of this paper, section III 

%\section{Contribution}

%This paper contributes to the field of demonstrating the utility of employing a tangent-based approach, instead of taking an angle-based approach for determining the curvature estimation for chord-based corner detectors. We propose a simple but effective technique which utilizes the acute angle between the two sides of a corner. This technique also contributes to better understanding of what a corner is and how it is detected. Therefore it also renders into better practical result.

\section{Background}\label{background}

A concise discussion of three chord-based detectors is presented in this section. These detectors are CPDA, CTAR, and CTAA. Since our work is inspired by useful insights into these methods, it is only appropriate that we discuss them in this section.

As CPDA is the predecessor of CTAR, and CTAA is the intellectual successor of both CPDA and CTAR, they all share some preprocessing steps. All three of them start with the detection of curves from images, similar to what most of the other chord-based detectors do \cite{PE:PG:PG2011short:037-042, awrangjeb2009fast, awrangjeb2008robust, sadat2011corner}. With the help of Canny edge detector, edges are obtained from planar curves. The edges then go through a smoothing process which utilizes a Gaussian kernel. After that, T junctions happen to be identified and then stored.

\begin{figure}[bh]
	\centerline{\includegraphics[width=68.7mm]{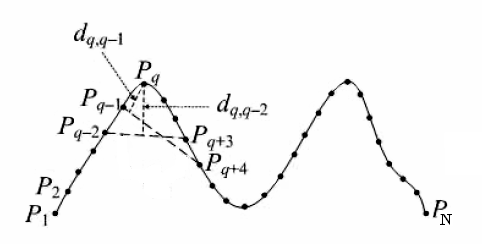}}
	\caption{Estimation of curvature using CPDA with chord \cite{awrangjeb2008robust}}
	\label{fig:cpda_process}
\end{figure}

Three discrete values of curvature are rendered for every location on extracted curves by CPDA. This is done by using three separate chords having lengths 10, 20, 30. Let each one of these chords be termed $L_i$ where i $\epsilon$ {10, 20, 30}. They get moved along each of the curves. In Fig. \ref{fig:cpda_process}, $N$ points on the curve are termed as $P_1$, $P_2$, $P_3$, $....$, $P_N$. Therefore, the chord $L_i$ is a straight line which joins the points $P_j$ and $P_{j+i}$ that are on the curve and are themselves $i$ points apart. For estimating the value of curvature $h_{L_i(q)}$ at the point $P_q$ using chord of length $i$, CPDA moves the chord on each side of $P_q$ for a maximum of $i$ points, while maintaining $P_q$ as interior point. Thus distance $d_{q, j}$ measuring the distance from $P_q$ to the chord is determined. Finally, the value of curvature is accumulated using (\ref{eq_dist_accumulation}).

\begin{equation}\label{eq_dist_accumulation}
   h_{L_i}(q)=\sum\limits_{j=q-i+1}^{q-1} {d_{q, j}}
\end{equation}

After that, the values of curvature that were estimated using every chord gets normalized using the normalizing equation mentioned in (\ref{eq_normalize}).

\begin{equation}\label{eq_normalize}
    h'{_{L_i}(q)}=\frac{h_{L_i}(q)}{max(h_{L_i})}, \mbox{$i \in \{ 10, 20, 30$\}, $1\leq q \leq N$}
\end{equation}

The values that were calculated for the three different chords using (\ref{eq_normalize}) are now multiplied together by using (\ref{eq_cpda_multiply}).

\begin{equation}\label{eq_cpda_multiply}
    H(q)=h'_{L_{10}}(q) \times h'_{L_{20}}(q) \times h'_{L_{30}}(q), \mbox{for $1\leq q \leq N$}
\end{equation}

Next, candidate corners are found out by rejecting the weak corners by the use of local maxima of the absolute curvature. It is done by comparing the values of curvature to threshold $T_h$. This threshold is set to $0.2$ by the authors of CPDA. Considering the hypothesis that ``a well defined corner should have a relatively sharp angle" \cite{he2004curvature} what CPDA does is calculating angle from a specific candidate corner to two of its neighboring candidate corners which are from previous step. This angle is then compared to the threshold angle $\delta$ with a view to removing false corners. The angle-threshold $\delta$ mentioned here is set to be $157^\circ$. In the end, CPDA compares T-junctions with detected corners and adds the T-junctions that are much further away from detected corners. This process of refinement is done with complex equations.

CTAR is another method which was first proposed in \cite{PE:PG:PG2011short:037-042}. In place of distance accumulation, another technique which has its roots in triangulation theory is used to determine the value of curvature. Each time a chord is put on curve, a brand new triangle can be generated by using two opposite ends of the chord along with the mid-point of segment of curve that is between two ends of above chord. The ratio of chord-length to summation of the length of another two sides of the triangle is taken. These two sides are defined to be from middle point of curve to two respective ends of that chord. The value of the aforementioned ratio is the estimated value of curvature for the mid-point of curve segment. As this measure is not using any kind of derivative based measurements, therefore it makes use of a relatively larger neighourhood.

%\begin{figure}[!t]
%\centering
%\includegraphics[width=2.5in]{Figures/ctar_process.png}
%\caption{Curvature estimation by using CTAR with chord  \cite{PE:PG:PG2011short:037-042}}
%\label{fig:ctar_process}
%\end{figure}

\begin{figure}[bh]
	\centerline{\includegraphics[width=68.7mm]{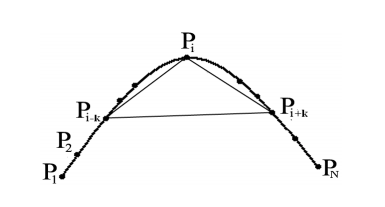}}
	\caption{Estimation of curvature using CTAR with chord \cite{PE:PG:PG2011short:037-042}}
	\label{fig:ctar_process}
\end{figure}

The above process will now be discussed with the help of an example depicted in Fig. \ref{fig:ctar_process}. Let $N$ points on a curve be $P_1$, $P_2$, $...$, $P_N$. Let $P_i$ be the point in which the value of curvature needs to be determined. First $k$ points from the point $P_i$ is traversed in forward direction leading to point $P_{i+k}$. Then again, $k$ points from $P_i$ in the backward direction is traversed leading to point $P_{i-k}$. If these three points, namely $P_{i-k}$, $P_i$, $P_{i+k}$ happen to be collinear, then the ratio of chord length between $P_{i-k}$ and $P_{i+k}$, to the sum of the length of the remaining two sides of this triangle, from $P_i$ to $P_{i-k}$ and from $P_i$ to $P_{i+k}$ in that order, will be $1$. Otherwise the ratio will be less that $1$. The value of this ratio is inversely proportional to sharpness of the corner at point $P_i$. Now, on point $P_i$, the value of curvature is determined using (\ref{CTAR_equation}).

\begin{equation}\label{CTAR_equation}
{R_L}(i)=\frac{d_1}{{d_2}+{d_3}}
\end{equation}

If the value of ${R_L}(i)$ is lower than the value of the threshold $T_h = 0.9896$, then $P_i$ gets considered as a candidate corner \cite{PE:PG:PG2011short:037-042}. In comparison, CTAR is faster than CPDA in terms of computation \cite{awrangjeb2013performance} and can find corners in a more reliable fashion under a range of image transformations \cite{PE:PG:PG2011short:037-042}.

Now we move on to the description of CTAA. Both chord-based detectors described so far in this section calculate values of curvature with help of length of sides of a number of triangles that are drawn around a specific point or a specific edge. CTAA simply calculates the angle between the sides of the triangle which originates from the point of edge \cite{DBLP:conf/icsipa/HossainMS15}.

%\begin{figure}[!t]
%	\centering{\includegraphics[width=68.7mm]{Figures/isosceles.png}}
%	\caption{Calculation of process CTAR threshold counterpart angle $\theta$ \cite{DBLP:conf/icsipa/HossainMS15}}
%	\label{fig:isosceles}
%\end{figure}

\begin{figure}[bh]
	\centerline{\includegraphics[width=68.7mm]{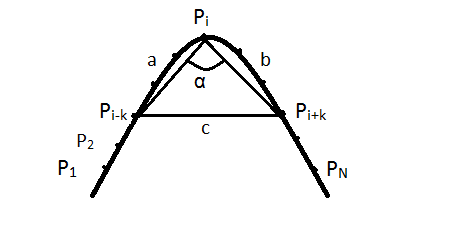}}
	\caption{Estimation of curvature using CTAA \cite{DBLP:conf/icsipa/HossainMS15}}
	\label{fig:proposed}
\end{figure}

CTAA is similar to CTAR and CPDA in the sense that all of these chord detectors traverse every single point on extracted curve. When CTAA detector is traversing the entire curve, it calculates the value of curvature for every point. Using neither accumulated distance method nor the ratio of length of sides, CTAA actually calculates angle $\alpha$ that is formed by this triangle in Fig. \ref{fig:proposed}. Let $P_1$, $P_2$, $P_3$, $...$, $P_n$ be a set of points called $P$ on this curve. $P_i$ is a point where the value of curvature needs to be calculated. Considering $k = 3$, the two other significant points are $P_{i-k}$ and $P_{i+k}$ \cite{DBLP:conf/icsipa/HossainMS15}. An acute angle $\alpha$ can now be calculated using (\ref{proposed_equation}).

\begin{equation}\label{proposed_equation}
\alpha = \cos^{-1}\frac{a^2+b^2-c^2}{2ab}
\end{equation}

Where, \newline\newline
\begin{equation}
a = \sqrt{(P_{ix}-P_{(i-k)x})^2+(P_{iy}-P_{(i-k)y})^2}
\end{equation} 
\begin{equation}
b = \sqrt{(P_{ix}-P_{(i+k)x})^2+(P_{iy}-P_{(i+k)y})^2} \end{equation} 
\begin{equation}
c = \sqrt{(P_{(i+k)x}-P_{(i-k)x})^2+(P_{(i+k)y}-P_{(i-k)y})^2}
\end{equation} 

If angle $\alpha$  is lower than a certain threshold angle $\theta \approx 163.5^{\circ}$, that point is termed as candidate corner. The final refinement process is mostly similar to that of CTAR, where every local minima coming from candidate corners gets considered as actual corners.

\section{Description of Proposed Technique}

\begin{figure}[bh]
	\centerline{\includegraphics[width=62mm]{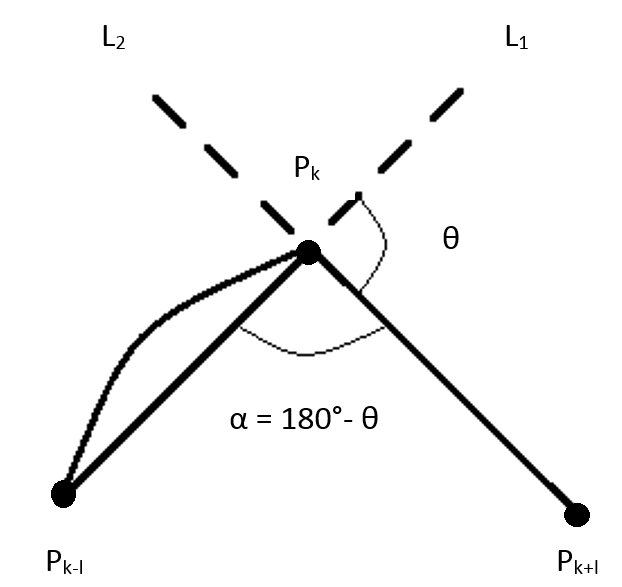}}
	\caption{Curvature estimation using CADT}
	\label{fig:tempfig}
\end{figure}

%Our proposed technique uses a simple tangent-based approach which utilizes the inherent property of any line that is called tangent. A tangent to a specific point on a plane curve is defined by Leibniz to be a line through two infinitely close points on the curve [Reference].

We name our proposed technique as Chord Angle Deviation using Tangent (CADT) detector. In this technique, a chord is moved along each curve in order to measure value of curvature at each point of that curve. This chord is used to calculate the tangents of two pairs of points on the curve, each pair including our target point where the value of curvature is to be calculated. Then the two tangent measures are used to find out the candidate corners.

The detailed explanation along with example of our proposed technique is as follows. Let $P_1$, $P_2$, $P_3$, $...$, $P_n$ be a set of $n$ points on the curve in Fig. \ref{fig:tempfig}. $P_k$ is the the middle point of the chord where we need to find the value of curvature. Hence the chord is stretching from point $P_{k-l}$ up to point $P_{k+l}$, thus the length of the chord is $2l+1$. Let the line stretching from point $P_{k-l}$ up to point $P_k$ be called $L_1$, and the line stretching from point $P_k$ up to point $P_{k+l}$ be called $L_2$. Therefore the intersection of these two lines is point $P_k$, and the equation of tangent for these two lines is given in (\ref{tangent_chord1}) and (\ref{tangent_chord2}).

%\begin{equation}\label{L_l}
%l = \frac{L}{2}   -ekhane-floor-dite-hobe
%%\floor*{\frac{x}{2}} < \frac{x}{2} < \ceil*{\frac{x}{2}}
%\end{equation} 

%we first traverse $k$ points from $P_k$ to the previous points up to point $P_{k-l}$ and thus get out first chord of length $l$ with one end at point $P_k$ and another at point $P_{k-l}$. This chord defines the tangent of the imaginary straight line stretching from point $P_{k-l}$ to point $P_k$. Next we traverse $k$ points from $P_k$ in the opposite direction up to point $P_{k+l}$, and thus get our second chord of length $l$ with one end at point $P_k$ and another at point $P_{k+l}$. This chord defines the tangent of the imaginary straight line stretching from point $P_{k}$ to point $P_{k+l}$. 

%Let $m_{a, b}$ define the tangent of the line whose two end points are $a$ and $b$ respectively. Let the coordinates for the points $a$ and $b$ be $\left( x_a, y_a\right) $ and $\left( x_b, y_b\right) $ respectively. The equation for measuring the tangent of this line is in (\ref{tangent_general}). 

%\begin{equation}\label{tangent_general}
%m_{a, b} = \frac{dy}{dx}
%\end{equation} 

%Here $dy$ denominates the difference of vertical co-ordinates of the two points $a$ and $b$, and $dy$ denominates the difference of horizontal co-ordinates of the two points $a$ and $b$. For finding out the tangent of our two chords, we use (\ref{tangent_chord1}) and (\ref{tangent_chord2}).

\begin{equation}\label{tangent_chord1}
m_{L_1} = \frac{y_{P_k} - y_{P_{k-l}}}
						{x_{P_k} - x_{P_{k-l}}}
\end{equation}

\begin{equation}\label{tangent_chord2}
m_{L_2} = \frac{y_{P_{k+l}} - y_{P_k}}
						  {x_{P_{k+l}} - x_{P_{k}}}
\end{equation}

Next, we compute the acute angle between these two lines. For this purpose we use the inverse tangent function as shown in (\ref{theta}).

\begin{equation}\label{theta}
\theta = \arctan
		\Biggl|
		\frac{m_{L_1} - m_{L_2}}
					 {1 + m_{L_1}m_{L_2}}
		\Biggr| 
\end{equation}

Now, the supplementary angle of $\theta$ is termed as $\alpha$, equation for which is expressed in (\ref{alpha}).

\begin{equation}\label{alpha}
\alpha = 180 ^{\circ} - \bigl| \theta \bigr|
\end{equation}

$\alpha$ is termed as the value of curvature of point $P_k$. If for any point this value of curvature is lower than the threshold angle of our proposed method $T_{h_{\alpha}} = 158.4^{\circ}$, then that point is considered as a candidate corner. For our proposed solution, we proved experimentally that a chord length of 9, i.e. $l = 4$ produces the best results. The refinement process of our proposed method is quite similar to the simple refinement process of CTAA. This process involves finding out all the local minima from the set of candidate corners and considering them as actual corners \cite{DBLP:conf/icsipa/HossainMS15}. 

%While CTAA and CTAR resorts to the use of the mathematical function of square root for calculating distance values for the triangle mentioned in Section \ref{background}, our proposed technique instead uses simple mathematical functions such as addition, subtraction, multiplication, and division. Therefore in terms of computational complexity, our proposed method is better than both CTAA and CTAR. ?????

%For calculating the value of curvature, CPDA, CTAR, CTAA, and our proposed detector use a range of mathematical operations. Each of the operations uses some significant as well as non-significant computational complexity. According to \cite{PE:PG:PG2011short:037-042}, square root operation is computationally the most expensive operation operation. All three detectors of CPDA, CTAR, and CTAA use the square root operation to calculate their respective values of curvature which depends on the Euclidean distance between two points on a plane. In contrast to the above mentioned detectors, our proposed detector does not use such complex and expensive mathematical operation. Therefore, we can say that our proposed detector computationally effective. 

\subsection{Image Accumulation and Transformation}
A grand total of 23 divergent images are used. The images are gray-scale images and include both artificial images as well as 
real world ones. The same collection of images were employed in the original work of CPDA and CTAR \cite{PE:PG:PG2011short:037-042, awrangjeb2008robust}. Maximum of these images are gathered from standard databases \cite{photo_db1, photo_db2}. Table \ref{table_database} demonstrates the transformations that are applied to this set of images in order to obtain a grand total of 8350 transformed test images. 

\begin{table}[!h]
\centering
\caption{Image Transformations applied on 23 base images \cite{DBLP:conf/icsipa/HossainMS15}}
\begin{small}
{\begin{tabular}{|M{2cm}|M{3cm}|M{2cm}|}
\hline
	\bfseries Transformations & \bfseries Transformation factors & \bfseries Number of images \\ \hline
	Scaling & Factors of scale $s_x$=$s_y$ in [0.5, 2.0] at .1 intervals, excluding 1.0 & 345 \\ \hline
	Shearing & Factors of shear $sh_x$ and $sh_y$ in [0, 0.012]at 0.002 intervals. & 1081 \\ \hline
	Rotation & 18 separate angles  of range $-90^\circ$ to $+90^\circ$ at $10^\circ$  & 437 \\ \hline
	Rotation-Scale & in [-30, +30 ] at $10^\circ$ intervals, followed by uniform and non uniform factors of scale $s_x$ and $s_y$ in [0.8, 1.2] at 0.1 apart. & 4025 \\ \hline
	Nonuniform Scale & Scale factors $s_x$ in [0.7, 1.3] and $s_y$ in [0.5, 1.5] at 0.1 intervals. & 1772 \\ \hline
	JPEG compression & Compression at 20 quality factors in [5, 100] at 5 intervals. & 460 \\ \hline
	Gaussian noise & Gaussian (G) noise at 10 variances in [0.005, 0.05] at 0.005 intervals. & 230 \\ \hline
\end{tabular}}{\label{table_database}}
\end{small}
\end{table}

\subsection{Evaluation Metrics}

Two oft-used metrics in the field of corner detection are average repeatability and localization error. These metrics are used for quantitative judgment of performance \cite{PE:PG:PG2011short:037-042, awrangjeb2012performance, awrangjeb2008robust, awrangjeb2007affine}. In terms of the set up of our experiment, average repeatability is simply a tool to measure the mathematical mean of the number of repeated corners that have been identified by corner detector in discussion between the original and the transformed images \cite{awrangjeb2008robust}. It has been defined as in (\ref{eval_metr}).

\begin{equation}\label{eval_metr}
Average Repeatability = 100\% \times \frac{\frac{A_p}{B_q}+
								 \frac{A_p}{C_r}}{2}
\end{equation}

A corner is termed as ``repeated" if it falls within a radius having 3 pixels of its expected location inside transformed image. This is the criteria that has been used by \cite{PE:PG:PG2011short:037-042, awrangjeb2008robust}, where $C_r$ and $B_q$ are respectively counts of corners that have been detected in the test and original images, and $A_p$ is the count of repeated corners between these two images. On the other hand, localization error is the measure of amount of pixel deviation of the repeated corners \cite{awrangjeb2008robust}. This can be measured by computing the root-mean-square-error, which is abbreviated as RMSE, of the locations of repeated corners in the test and original images. Equation of localization error is as in (\ref{RMSE}),

\begin{equation}\label{RMSE}
L_e=\sqrt{\frac{1}{A_p}\sum\limits_{i=1}^{A_p}(x_{mi}-x_{ni})^2 + (y_{mi}-y_{ni})^2}
\end{equation}

where $\left(x_{mi}, y_{mi}\right)$ and $\left(x_{ni}, y_{ni}\right)$ are positions of $i$-th repeated corner of respectively the original and test images.

\subsection{Results and Discussion}

\begin{figure}[bh]
	\centerline{\includegraphics[width=68.7mm]{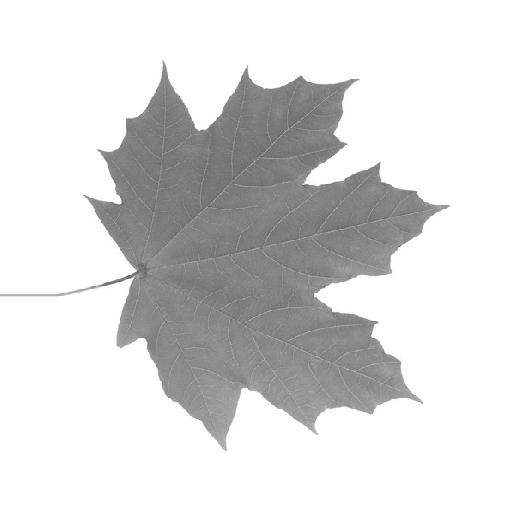}}
	\caption{Test image contained in experiment database \cite{DBLP:conf/icsipa/HossainMS15}}
	\label{fig:leaf}
\end{figure}

\begin{figure}[!t]
	\centering{\includegraphics[width=68.7mm]{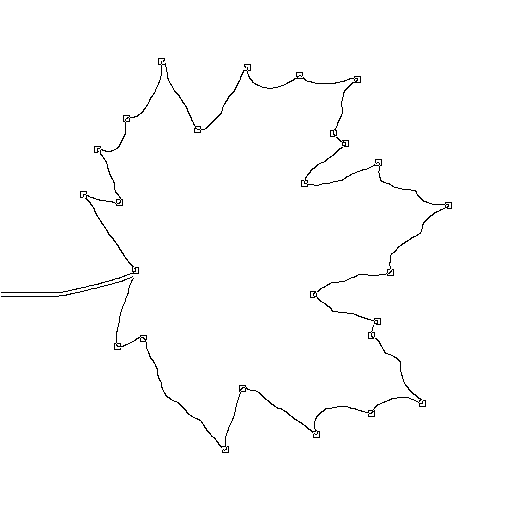}}
	\caption{Corners detected by CPDA process \cite{DBLP:conf/icsipa/HossainMS15}}
	\label{fig:CPDA}
\end{figure}

\begin{figure}[!t]
	\centering{\includegraphics[width=68.7mm]{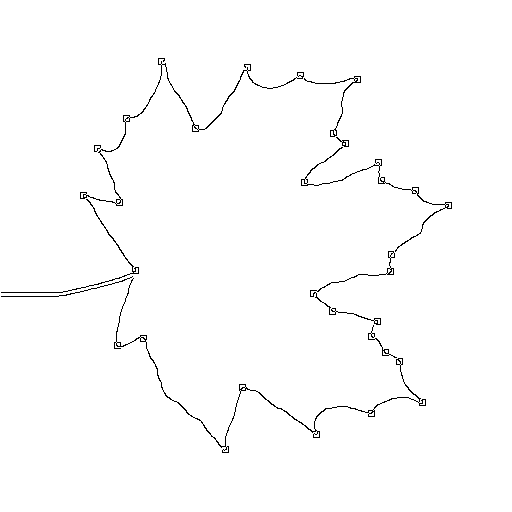}}
	\caption{Corners detected by CTAR process \cite{DBLP:conf/icsipa/HossainMS15}}
	\label{fig:CTAR}
\end{figure}

\begin{figure}[!t]
	\centering{\includegraphics[width=68.7mm]{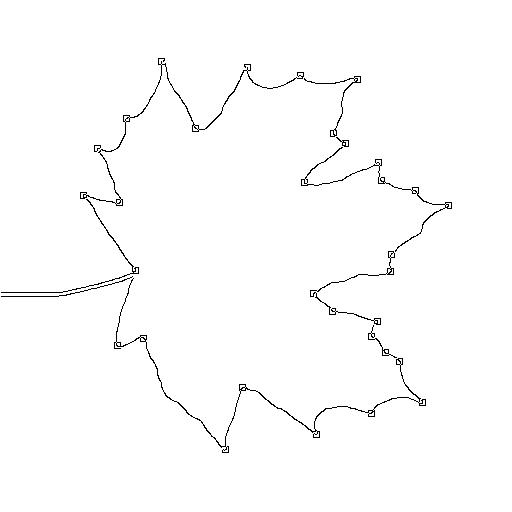}}
	\caption{Corners detected by CTAA process \cite{DBLP:conf/icsipa/HossainMS15}}
	\label{fig:CTAA}
\end{figure}

\begin{figure}[!t]
	\centering{\includegraphics[width=68.7mm]{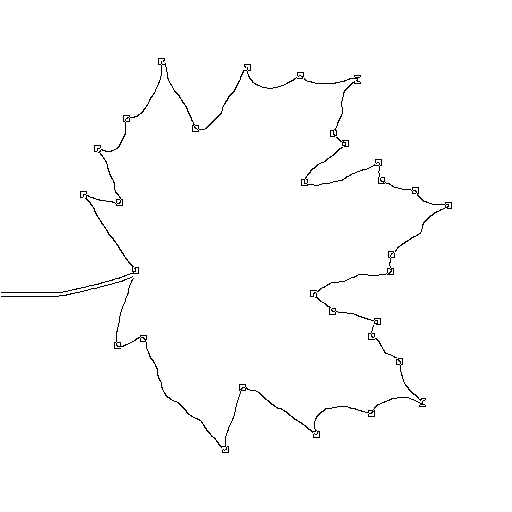}}
	\caption{Corners detected by CADT process}
	\label{fig:Tangent}
\end{figure}

In Fig. \ref{fig:leaf} we show one image of a leaf from the test image database that has been used in our experiment. Fig. \ref{fig:CPDA}, \ref{fig:CTAR}, \ref{fig:CTAA}, and \ref{fig:Tangent} depict the corners that have been detected from the leaf image by CPDA, CTAR, CTAA, and CADT respectively.

\begin{figure}[!t]
	\centering{\includegraphics[width=77mm]{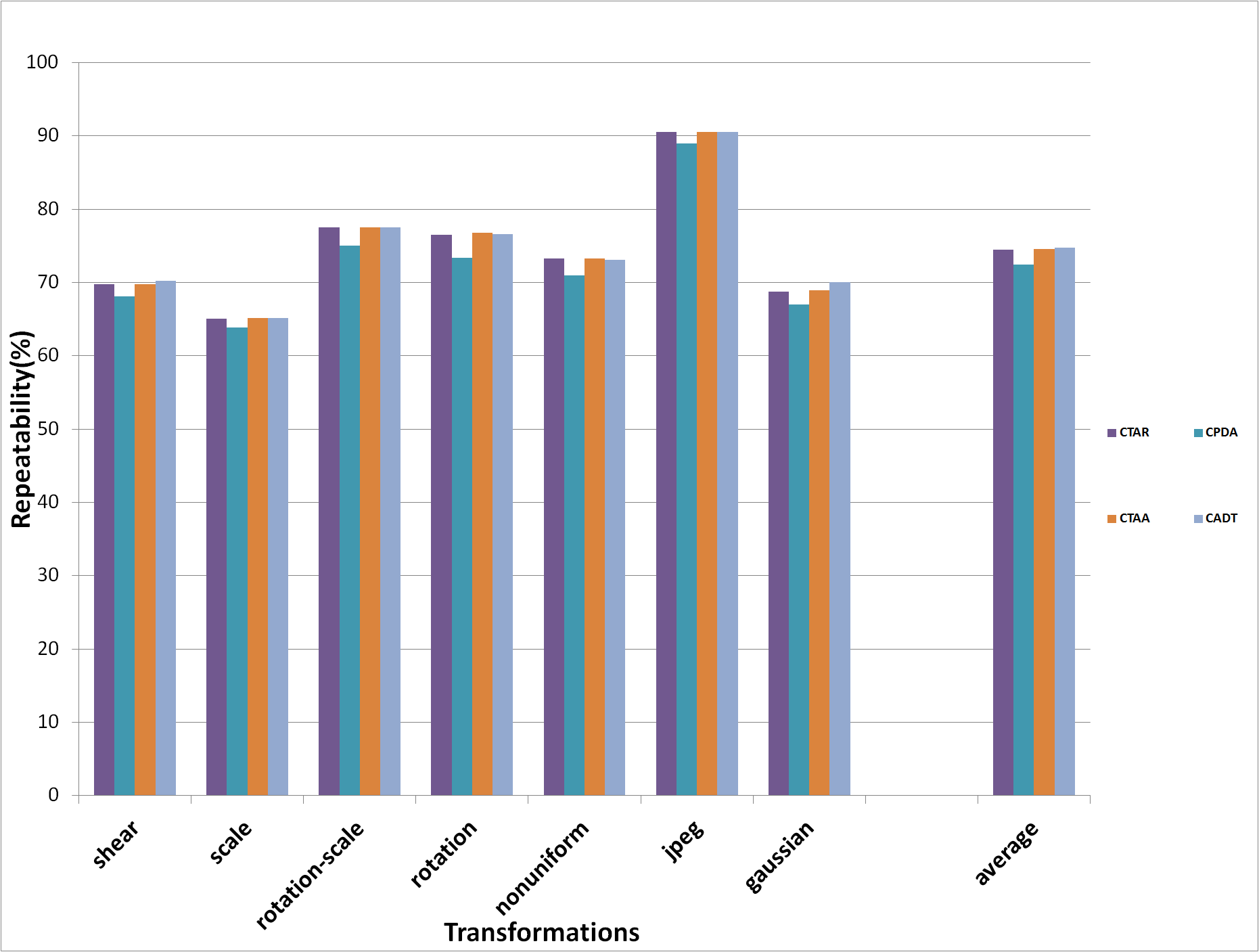}}
	\caption{Average repeatability of corner detectors on various transformed images}
	\label{fig:Repeatability}
\end{figure}

Fig. \ref{fig:Repeatability} depicts a comparative indication for the metric of average repeatability of CPDA, CTAR, CTAA, and CADT for seven distinct transformations apiece, as well as the mean of average repeatability across all the transformations. From this comparative picture, we see that in terms of average repeatability, tangent process outperforms CTAA, which in turn outperforms CTAR as well as CPDA. This comparison signifies that by adopting our proposed approach, more reliable corner detection can be achieved, in comparison to using the ratio of length of triangle sides, or accumulated distance (as with CPDA).

Table \ref{tab:basic} demonstrates the count of corners that have been detected from the original images and the average repeatability (across all transformations) of these corner detectors. 

\begin{table}[!h]
\caption{Average repeatability and number of corners detected by corner detectors}

\centering
{\begin{tabular}{|c|c|c|}
\hline
	\bfseries Methods &  \bfseries Average repeatability & \bfseries Corner count \\  \hline 
	CPDA & 72.5 & 882 \\ 
	CTAR & 74.48 & 1364 \\ 
	CTAA & 74.56 & 1367 \\ 
	CADT & 74.77 & 1351 \\ \hline
\end{tabular}}{\label{tab:basic}}
\end{table}

%another table needed, ke kar moddhe koto bar (corner) ase

\begin{figure}[!t]
	\centering{\includegraphics[width=90mm]{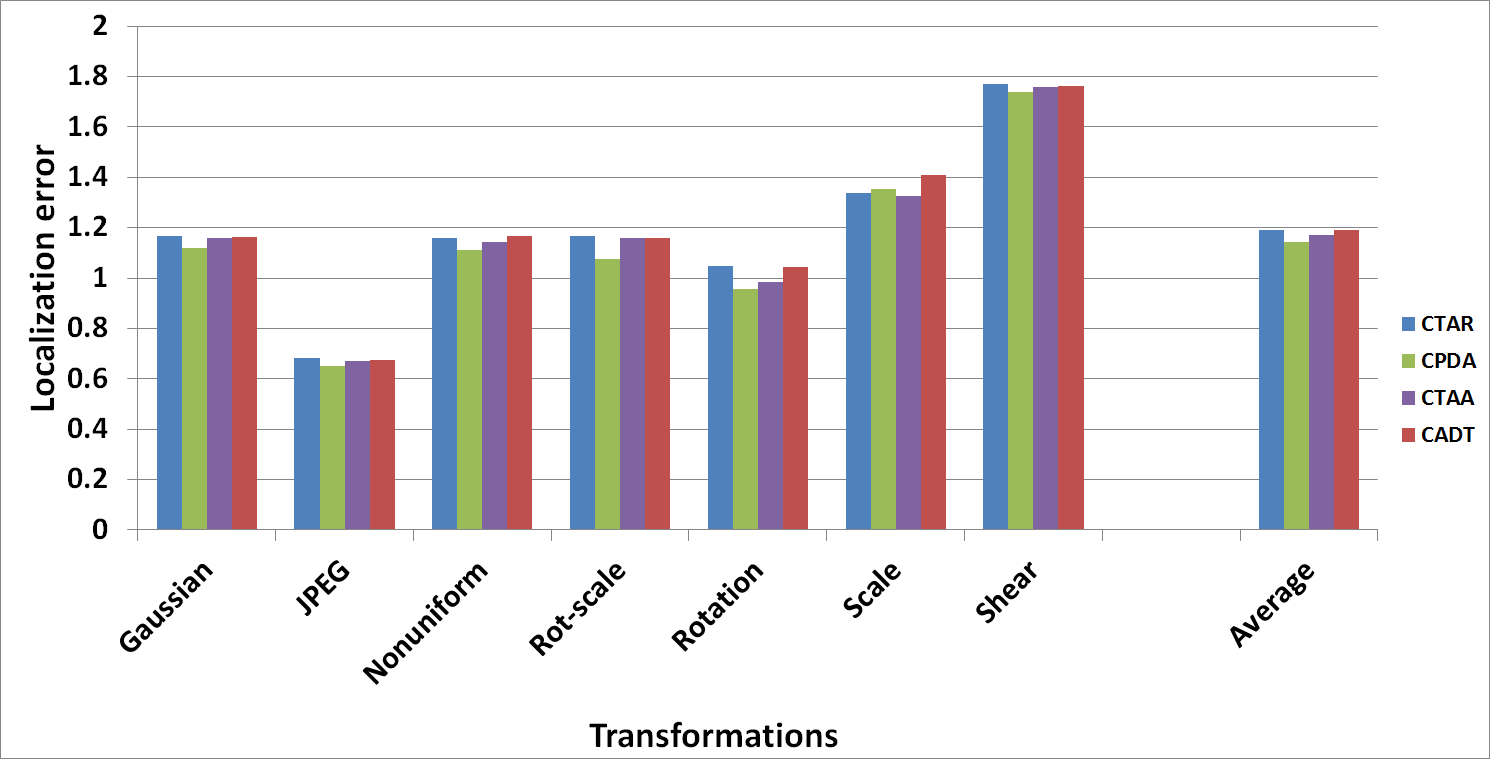}}
	\caption{Localization of error of corner detectors on various transformed images}
	\label{fig:Localization}
\end{figure}

Fig. \ref{fig:Localization} depicts a comparative indication for the metric of average repeatability of CPDA, CTAR, CTAA, and CADT. CADT and CTAR have identical localization error value 2hich is approximately 1.19. On the other hand, CADT has slightly higher localization error value than that of CTAA, which can be a topic for future research for performance improvement. 

For calculating the value of curvature, CPDA, CTAR, CTAA, and CADT use a range of mathematical operations. Each of the operations uses some significant as well as non-significant computational complexity. According to \cite{PE:PG:PG2011short:037-042}, square root operation is computationally the most expensive operation. All three detectors of CPDA, CTAR, and CTAA use the square root operation to calculate their respective values of curvature which depends on the Euclidean distance between two points on a plane. In contrast to the above mentioned detectors, our proposed detector does not use such complex and expensive mathematical operation. Therefore, we can say that our proposed detector is computationally effective.

%\newpage
\section{Conclusion}

%We have proposed an efficient and robust new method for corner detection of an image in this paper. Our novel approach of using difference of tangent values of chords as value of curvature has been compared against the estimated corner angle-based calculations of CTAA as well as triangular arm length-based calculations of CTAR and CPDA. We have adopted the approach of using a fixed threshold angle a fixed chord length. This paper also exhibited that experimentally tangent-based approach performs better than CTAA in the field of average repeatability.

Our paper focuses at proposing an effective and reliable corner detector to identify corners in an image. We have used a set of simple mathematical calculations to estimate the value of curvature. We have compared our experimental results with CTAA, CTAR, and CPDA, where we have found that our proposed corner detector achieves the best average repeatability in detecting strong corners. Nonetheless the localization error of our proposed detector is slightly higher than that of CPDA and same as that of CTAR. Our proposed method is computationally faster and efficient.

\IEEEtriggeratref{6}
\bibliography{Bib/bib}{}
\bibliographystyle{IEEEtran}

%\end{thebibliography}

\end{document}